\documentclass[11pt]{article}

\usepackage[utf8]{inputenc}
\usepackage[T1]{fontenc}
\usepackage{lmodern}
\usepackage{microtype}
\usepackage{placeins}

\usepackage[margin=1in]{geometry}
\usepackage{setspace}
\usepackage{parskip}

\usepackage{amsmath}
\usepackage{amssymb}
\usepackage{graphicx}
\usepackage{booktabs}
\usepackage{array}
\usepackage{multirow}

\usepackage{url}
\usepackage{hyperref}
\usepackage{xcolor}

\hypersetup{
    colorlinks=true,
    linkcolor=black,
    citecolor=black,
    urlcolor=blue,
    pdftitle={NeuroSynth: A Biologically Inspired Continual Reinforcement Learning Architecture for Mitigating Catastrophic Forgetting},
    pdfauthor={Yash Kini}
}

\title{
    \textbf{NeuroSynth: A Biologically Inspired Continual Reinforcement Learning Architecture for Mitigating Catastrophic Forgetting}
}

\author{
    Yash Kini\\
    James Madison High School\\
    Vienna, Virginia, USA\\
    \href{mailto:kiniyash3@gmail.com}{kiniyash3@gmail.com}
}

\date{}

\begin{document}

\maketitle

\begin{abstract}
Artificial Intelligence (AI) systems often perform well on isolated tasks but struggle under continual learning conditions, where training on new tasks can overwrite previously acquired knowledge, a failure mode known as catastrophic forgetting. Biological learning systems reduce this interference through complementary memory processes involving rapid hippocampal encoding and slower cortical consolidation. This study introduces NeuroSynth, a brain-inspired continual reinforcement learning architecture designed to mitigate catastrophic forgetting through a dual-pathway consolidation mechanism. NeuroSynth separates rapid task acquisition from long-term retention using distinct ``plan'' and ``habit'' pathways combined with replay and knowledge distillation. NeuroSynth was evaluated against Proximal Policy Optimization (PPO) and Elastic Weight Consolidation (EWC) across three sequential navigation tasks with changing goal locations in a non-revisitation continual learning setting. Across six independent seeds, NeuroSynth preserved substantially more early-task knowledge than PPO after sequential training, achieving 18.00\% Task A success rate compared to 0.33\% for PPO ($p = 0.014929$, Cohen's $d = 1.49$) and 35.33\% Task B success rate compared to 0.00\% for PPO ($p = 0.002376$, Cohen's $d = 2.31$). NeuroSynth also demonstrated higher final Task C performance than EWC, achieving 9.00\% compared to 2.00\% ($p = 0.226643$, Cohen's $d = 0.56$), indicating a moderate but not statistically significant advantage. These findings suggest that biologically inspired consolidation mechanisms may improve the stability-plasticity balance in continual reinforcement learning systems.
\end{abstract}


\section{Introduction}

Catastrophic forgetting occurs when neural networks lose performance on previously learned tasks after sequential training on new tasks because gradient updates overwrite parameters that were important for earlier behaviors (1, 2). This phenomenon remains a major limitation in continual reinforcement learning because most reinforcement learning architectures are optimized under stationary task assumptions rather than sequential adaptation conditions (3, 4, 15). When task objectives change over time, updates that improve performance on new tasks frequently overwrite parameters associated with earlier behaviors, causing rapid degradation of prior knowledge. In reinforcement learning, catastrophic forgetting has been observed across policy-gradient and value-based methods, including Proximal Policy Optimization (PPO) and Deep Q-Network (DQN) architectures (3, 5, 12). This limitation constrains the deployment of autonomous systems operating in dynamic environments requiring continual adaptation, including robotics, autonomous navigation, and medical artificial intelligence systems (6).

Several continual learning methods have been proposed to reduce catastrophic forgetting. Replay-based methods mitigate forgetting by storing and revisiting prior experiences during future optimization steps (4, 17, 18). Although replay stabilizes learning, sequential gradient updates may still overwrite internal representations associated with earlier tasks. Regularization-based approaches such as Elastic Weight Consolidation (EWC) attempt to preserve important parameters by penalizing updates to weights estimated to be critical for prior tasks using Fisher Information matrices (10). While EWC reduces catastrophic forgetting in some environments, strong parameter constraints may impair adaptation to later tasks under highly sequential learning conditions. Because all tasks share a single parameter space, preserving earlier weights may increasingly restrict representational flexibility as new tasks accumulate. Progressive Neural Networks reduce interference by freezing prior networks and adding new subnetworks for future tasks (11). However, these approaches require continual model expansion and do not perform representational consolidation within a unified memory system.

In contrast to artificial neural networks, biological learning systems demonstrate the ability to continuously acquire new information while preserving previously consolidated knowledge. Complementary Learning Systems (CLS) theory proposes that this capability emerges through interactions between the hippocampus and neocortex (12). The hippocampus rapidly encodes episodic information, while cortical systems gradually consolidate stable long-term representations over extended timescales (12, 13). Experimental evidence suggests that replay and memory reactivation during sleep may contribute to stabilization and long-term consolidation of prior experiences (14, 15).

Despite strong biological evidence supporting complementary learning systems, most artificial continual learning architectures primarily focus on parameter protection or data rehearsal rather than explicit memory-system separation (5, 6, 15). Existing reinforcement learning approaches generally maintain a single adaptive representation space, making previously learned behaviors vulnerable to being overwritten as new tasks are acquired. This work explores the design and evaluation of NeuroSynth, a brain-inspired continual reinforcement learning architecture that translates complementary hippocampal-cortical consolidation principles into a functional algorithmic framework. NeuroSynth introduces architectural separation between rapid acquisition and slower consolidation pathways through dual-network memory organization, replay-based rehearsal, and policy distillation mechanisms. Specifically, this study evaluates the performance of NeuroSynth, a biologically inspired continual reinforcement learning architecture, under sequential non-revisitation learning conditions.

NeuroSynth was evaluated against two widely used continual reinforcement learning baselines: Proximal Policy Optimization (PPO) and Elastic Weight Consolidation (EWC). All models were trained sequentially on three navigation tasks within the same environment but with different goal locations. Once a task phase ended, it was never revisited during future training. This non-revisitation setting was specifically designed to induce catastrophic forgetting by forcing continual adaptation without direct retraining on prior tasks. Across six independent experimental seeds, NeuroSynth achieved 18.00\% Task A success rate compared to 0.33\% for PPO ($p = 0.014929$, Cohen's $d = 1.49$), and 35.33\% Task B success rate compared to 0.00\% for PPO ($p = 0.002376$, Cohen's $d = 2.31$). NeuroSynth also demonstrated improved final Task C performance relative to EWC, achieving 9.00\% versus 2.00\% ($p = 0.226643$, Cohen's $d = 0.56$), indicating a moderate but not statistically significant advantage. These findings suggest that biologically inspired consolidation mechanisms may improve the stability-plasticity balance in continual reinforcement learning systems.

\section{Materials and Methods}

\subsection{Experimental Design}

NeuroSynth was compared against two baseline reinforcement learning approaches:

\begin{itemize}
    \item Proximal Policy Optimization (PPO, 7)
    \item Elastic Weight Consolidation (EWC, 10)
\end{itemize}

All methods were trained sequentially on three navigation tasks. The environment structure remained constant across all tasks, while goal locations changed between phases. Once a task was completed, it was never revisited during future training. This sequential non-revisitation setting was designed to isolate catastrophic forgetting under controlled continual learning conditions.

To ensure experimental reliability, all methods were evaluated across six independent random seeds. Each method used identical environment configurations, training schedules, evaluation procedures, and observation preprocessing pipelines. Performance differences were therefore attributable to algorithmic design rather than environmental variation or initialization artifacts.

\subsection{NeuroMaze-CL Environment}

\begin{figure}[ht]
    \centering
    \includegraphics[width=\textwidth]{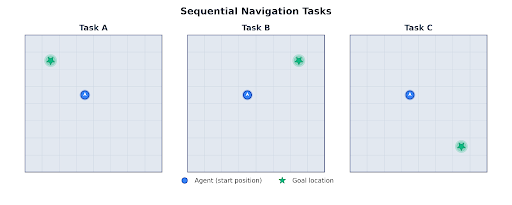}
    \caption{NeuroMaze-CL Sequential Navigation Tasks. Three navigation tasks trained sequentially in the order Task A $\rightarrow$ Task B $\rightarrow$ Task C. Goal locations change between tasks while the environment structure remains constant.}
    \label{fig:tasks}
\end{figure}

Experiments were conducted using NeuroMaze-CL, a custom continual reinforcement learning environment implemented using the Gymnasium framework. The environment consisted of an $8 \times 8$ grid-world navigation task in which agents learned to reach task-specific goal locations. The three sequential task configurations are illustrated in Figure~\ref{fig:tasks}.

\begin{figure}[ht]
    \centering
    \includegraphics[width=0.75\textwidth]{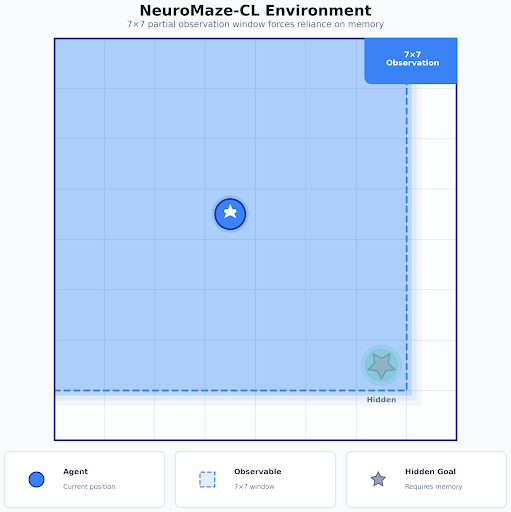}
    \caption{NeuroMaze-CL Partial Observability. The agent receives a $7\times7$ egocentric RGB observation window centered on its current position. Goals outside the observation window are not directly visible, requiring learned internal representations for navigation.}
    \label{fig:partial}
\end{figure}

Agents received partially observable egocentric RGB observations represented as $7 \times 7$ windows centered around the agent position. Observations were converted into normalized \texttt{float32} tensors with dimensions $3 \times 7 \times 7$. Partial observability reduced complete environmental visibility and increased dependence on learned internal representations rather than purely reactive policies, as illustrated in Figure~\ref{fig:partial}.

Each episode terminated when:

\begin{itemize}
    \item the agent reached the goal state, or
    \item the maximum episode step limit was reached.
\end{itemize}

Agents received a terminal reward of +1.0 upon reaching the correct goal state. Decoy goals terminated the episode with a reward of 0.0, while conflict decoys used during Task C produced a terminal reward of $-1.0$ to increase interference pressure against previously learned goal representations. During non-terminal transitions, agents received a step penalty of $-0.01$ combined with potential-based reward shaping proportional to reductions in Manhattan distance from the active goal state.

The maximum episode length was fixed at 150 steps across all experiments. The action space consisted of four discrete movement actions corresponding to cardinal navigation directions.

\subsection{Variables}

Table~\ref{tab:variables} summarizes the experimental variables. Success rate was defined as the percentage of evaluation episodes in which the agent successfully reached the goal state before episode termination.

\begin{table}[ht]
\centering
\caption{Experimental Variables}
\label{tab:variables}

\begin{tabular}{p{0.28\textwidth}p{0.62\textwidth}}
\toprule
\textbf{Variable Type} & \textbf{Variables} \\
\midrule

Independent Variables &
\begin{tabular}[t]{@{}l@{}}
Proximal Policy Optimization (PPO)\\
Elastic Weight Consolidation (EWC)\\
NeuroSynth
\end{tabular}
\\[0.6em]

Dependent Variables &
\begin{tabular}[t]{@{}l@{}}
Task Success Rates (\%) of Tasks A, B and C\\
Retention Metrics of Task A and B after Learning Task C
\end{tabular}
\\[0.6em]

Controlled Variables &
\begin{tabular}[t]{@{}l@{}}
Maze Structure\\
Observation Format\\
Reward Function\\
Training Steps\\
Evaluation Protocol
\end{tabular}
\\

\bottomrule
\end{tabular}
\end{table}

\subsection{PPO Baseline}

The PPO baseline implemented a standard actor-critic architecture based on Proximal Policy Optimization (7) using reinforcement learning principles described by Sutton and Barto (16). Training used clipped surrogate policy objectives, generalized advantage estimation, entropy regularization, and gradient clipping. PPO intentionally contained no continual learning mechanisms, replay systems, or task conditioning strategies. The PPO baseline therefore served as a catastrophic forgetting reference condition under sequential continual learning.

\subsection{EWC Baseline}

The EWC baseline implemented a double DQN value-based reinforcement learning architecture using Fisher Information regularization to constrain updates to parameters estimated to be important for prior tasks (10). Fisher matrices were computed using temporal-difference loss gradients sampled from task-specific replay buffers. Parameter penalties were averaged across prior tasks to reduce instability caused by excessive regularization scaling. This constrained updates to parameters estimated to contribute strongly to prior-task performance.

Unlike replay-based continual learning methods, EWC attempted to preserve prior knowledge primarily through parameter protection mechanisms rather than explicit representational rehearsal.

\subsection{NeuroSynth Architecture}

\begin{figure}[ht]
    \centering
    \includegraphics[width=\textwidth]{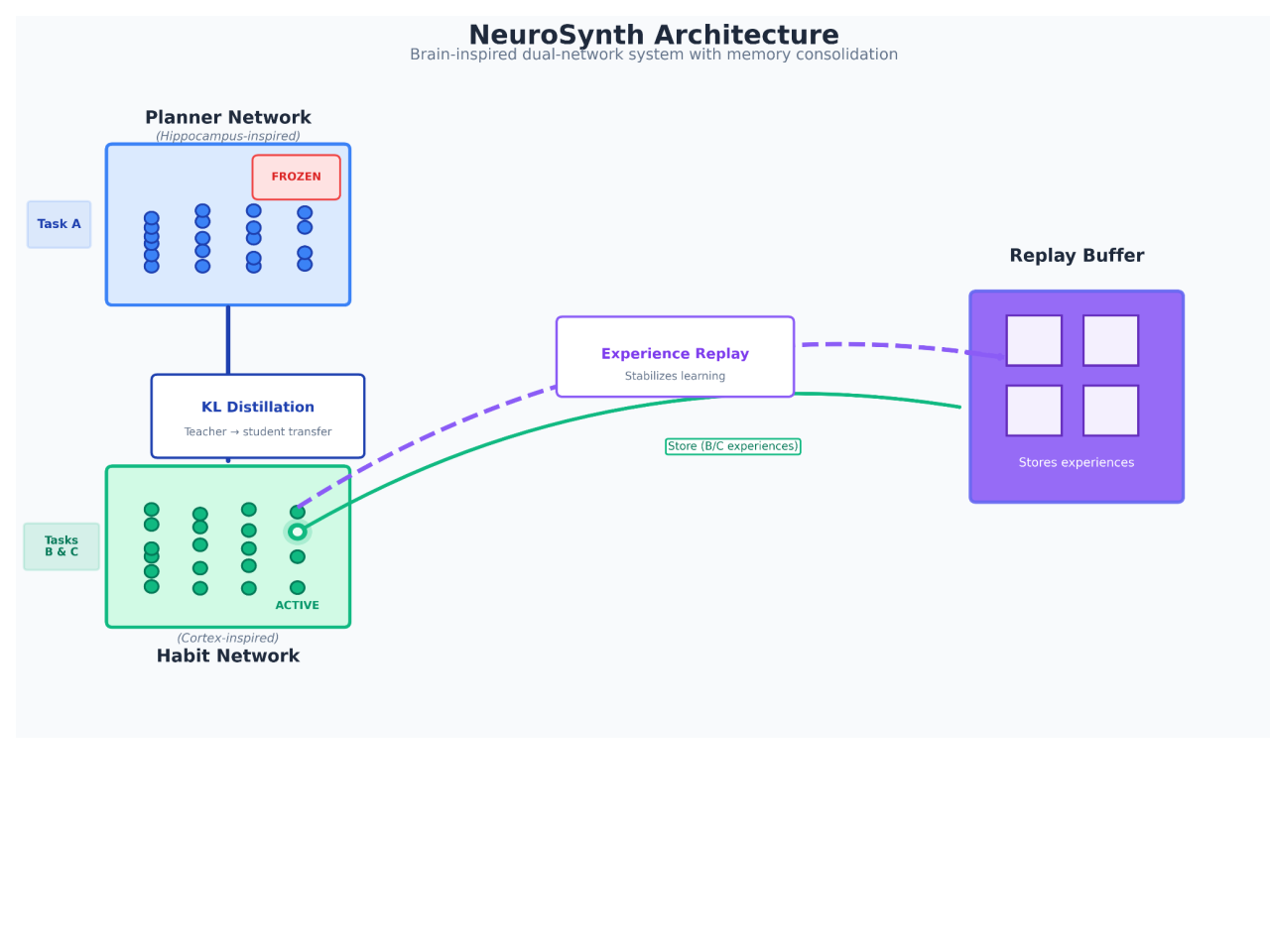}
    \caption{NeuroSynth Dual-Pathway Architecture. The hippocampal-inspired planner network rapidly acquires Task A representations and is subsequently frozen. The cortical-inspired habit network remains plastic during Tasks B and C, receiving behavioral guidance from the frozen planner via KL divergence distillation on replayed Task A states.}
    \label{fig:architecture}
\end{figure}

NeuroSynth implemented a dual-pathway continual reinforcement learning architecture inspired by complementary learning systems theories in neuroscience. The overall architecture is illustrated in Figure~\ref{fig:architecture}. The architecture separated rapid acquisition from slower consolidation through two complementary neural pathways:

\begin{itemize}
    \item A hippocampal-inspired ``plan'' pathway
    \item A cortical-inspired ``habit'' pathway
\end{itemize}

The plan pathway rapidly acquired Task A representations through temporal-difference learning. After Task A training, all planning parameters were frozen to preserve early representations. An exponential moving average (EMA) copy of the planning system was maintained as a teacher network during subsequent phases.

The habit pathway remained plastic during Tasks B and C. This pathway learned new tasks while preserving earlier knowledge through replay and policy distillation mechanisms.

\subsection{Replay Consolidation}

Replay mechanisms re-exposed the habit pathway to Task A state distributions during future learning phases (4). Replay buffers used reservoir sampling to reduce early-training bias and maintain representative historical samples across training.

Replay batches combined:

\begin{itemize}
    \item current-task experiences,
    \item and replayed prior-task experiences.
\end{itemize}

This design preserved exposure to earlier task distributions while allowing continued adaptation to new tasks.

\subsection{Policy Distillation}

Knowledge distillation aligned the adaptive habit pathway with the frozen planning pathway using Kullback--Leibler (KL) divergence losses with temperature scaling (17). Distillation was applied specifically on replayed Task A states.

The frozen plan pathway acted as a teacher model, while the habit pathway acted as a student model. Replay preserved exposure to earlier state distributions, while distillation stabilized policy behavior associated with those states. This process transferred stable behavioral representations without directly retraining the frozen planner.

\subsection{Training Procedure}

Training followed a fixed sequential order (see Figure~\ref{fig:tasks}):

\begin{itemize}
    \item Task A: Initial navigation objective.
    \item Task B: Second navigation objective with a distinct goal location.
    \item Task C: Final navigation objective with a third goal location.
\end{itemize}

No parameter resets occurred between tasks.

Training step budgets were:

\begin{itemize}
    \item Task A: 100,000 steps
    \item Task B: 100,000 steps
    \item Task C: 200,000 steps
\end{itemize}

Task C used a longer training duration to increase representational interference and strengthen catastrophic forgetting pressure.

Evaluation used deterministic seed schedules shared across PPO, EWC, and NeuroSynth to ensure direct comparability between methods. Each evaluation phase consisted of 50 episodes per task. Hyperparameter settings for all methods are summarized in Table~\ref{tab:hyperparameters}.

\begin{table}[ht]
    \centering
    \includegraphics[width=\textwidth]{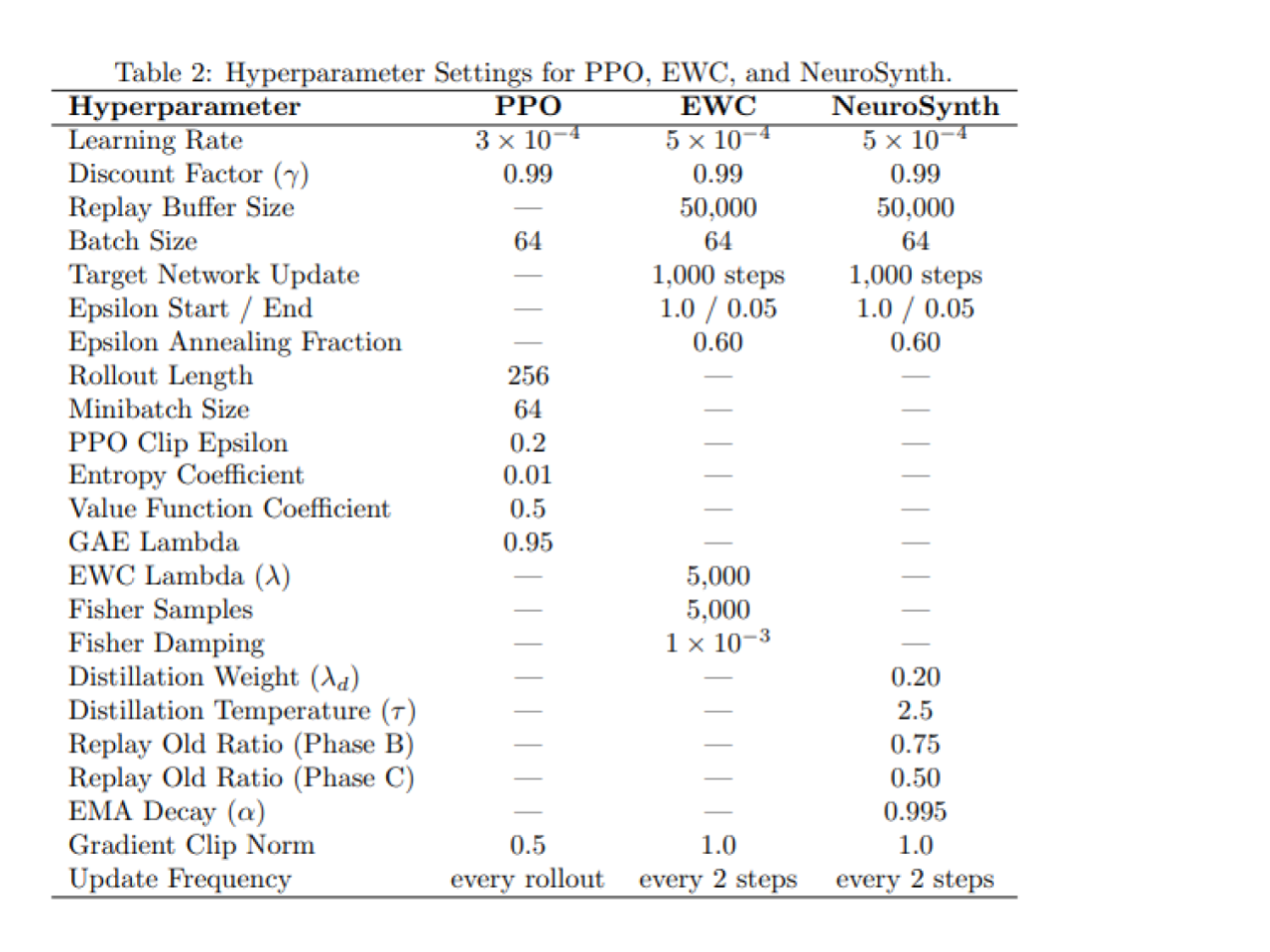}
    \caption{Hyperparameter settings for PPO, EWC, and NeuroSynth.}
    \label{tab:hyperparameters}
\end{table}

\subsection{Software and Hardware}

All experiments were implemented in Python 3.10 using PyTorch for neural network modeling (18). PPO implementations used Stable-Baselines3 components where applicable (21). NeuroMaze-CL was developed using the Gymnasium interface. Numerical processing used NumPy, and visualization generation used Matplotlib. Experiments were conducted on an NVIDIA RTX 4060 GPU.

Training reproducibility was enforced through:

\begin{itemize}
    \item deterministic seed initialization,
    \item deterministic evaluation schedules,
    \item synchronized environment resets,
    \item and deterministic PyTorch configurations where supported.
\end{itemize}

The complete source code, NeuroMaze-CL benchmark environment, training scripts, evaluation utilities, and statistical analysis scripts are publicly available at:

\begin{center}
\url{https://github.com/yashkini1/NEUROSYNTH}
\end{center}

\section{Results}

NeuroSynth was evaluated under sequential continual reinforcement learning conditions using three navigation tasks trained in the order Task A $\rightarrow$ Task B $\rightarrow$ Task C. Once training on a task was completed, the agent no longer revisited that task during future learning phases. This experimental design was intended to induce catastrophic forgetting by forcing each model to continually adapt without direct retraining on earlier tasks.

Performance was compared against two baseline methods:

\begin{itemize}
    \item Proximal Policy Optimization (PPO)
    \item Elastic Weight Consolidation (EWC)
\end{itemize}

All models were evaluated across six independent experimental seeds using standardized deterministic evaluation schedules.

\subsection{Final Task Performance}

\begin{figure}[ht]
    \centering
    \includegraphics[width=\textwidth]{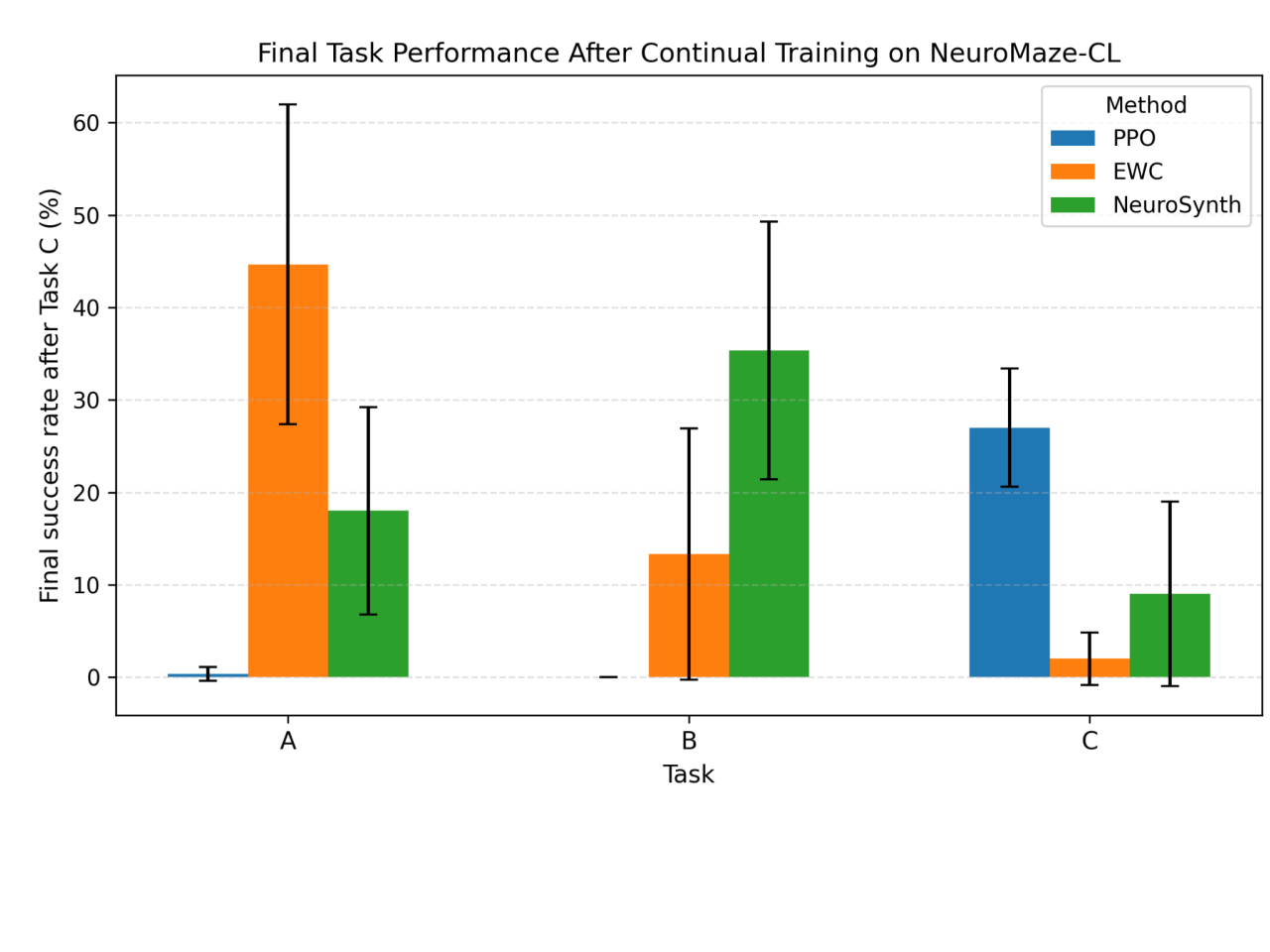}
    \caption{Final Task Performance After Continual Training on NeuroMaze-CL. Mean success rates (\%) across six independent seeds for PPO, EWC, and NeuroSynth on Tasks A, B, and C following sequential training. Error bars indicate one standard deviation.}
    \label{fig:final_performance}
\end{figure}

Figure~\ref{fig:final_performance} summarizes final task performance after the complete continual-learning sequence. Bars represent the mean success rate across six experimental seeds for each method on Tasks A, B, and C following sequential training. Error bars represent $\pm1$ standard deviation across seeds.

The PPO baseline demonstrated severe catastrophic forgetting. Although PPO was capable of learning newly introduced tasks, performance on earlier tasks collapsed after later phases of training. Final retention on Task A approached zero after Task C training, indicating near-complete overwriting of earlier representations.

EWC partially mitigated forgetting through parameter regularization. However, EWC demonstrated reduced adaptability during later tasks, particularly under Task C conditions. This suggests that Fisher-information constraints may preserve older representations at the cost of reduced plasticity for newly introduced tasks.

NeuroSynth achieved improved balance between stability and adaptability. Compared to PPO, NeuroSynth preserved substantially more Task A and Task B knowledge after sequential training while still maintaining the ability to acquire Task C. These findings are consistent with the hypothesis that the NeuroSynth architecture may improve continual learning stability relative to the evaluated baselines.

\subsection{Task A Retention Across Continual Learning Phases}

\begin{figure}[ht]
    \centering
    \includegraphics[width=\textwidth]{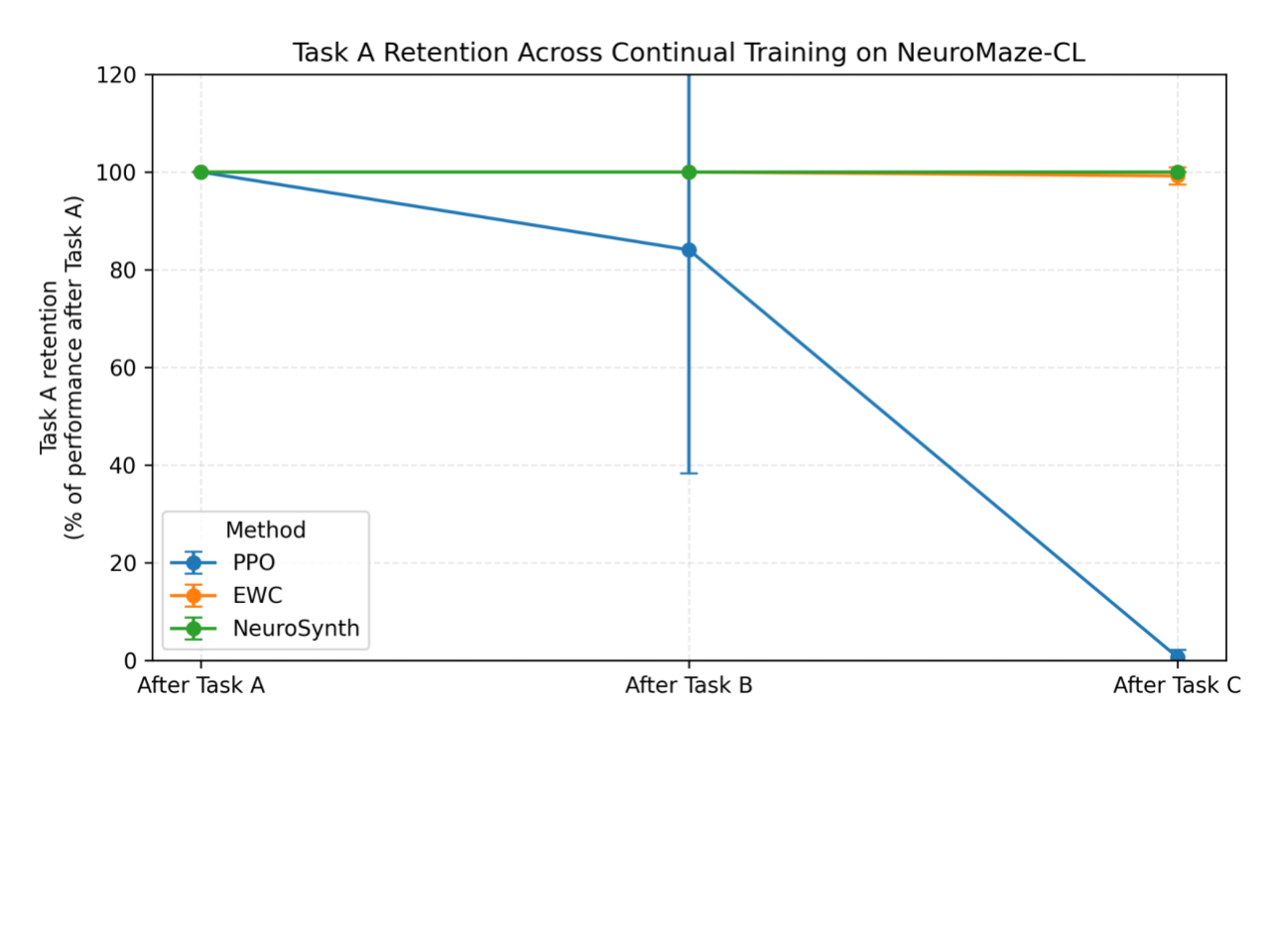}
    \caption{Task A Retention Across Continual Learning Phases. Task A success rate expressed as a percentage of post-Task A baseline performance, measured after each training phase. Points represent the mean success rate across six random seeds. Error bars denote $\pm1$ standard deviation.}
    \label{fig:taskA}
\end{figure}

Figure~\ref{fig:taskA} visualizes Task A retention measured after completion of Tasks A, B, and C. Error bars indicate $\pm1$ standard deviation across six experimental seeds. Following Task A training, all methods achieved their respective peak Task A performance. However, substantial divergence emerged after subsequent training phases.

PPO exhibited progressive degradation after Task B and severe collapse after Task C. Final Task A retention approached 0\%, demonstrating catastrophic forgetting under unrestricted gradient updates.

EWC maintained relatively stable Task A retention throughout continual training. Because EWC constrains updates to parameters estimated to be important for earlier tasks, previously learned representations remained comparatively protected from overwriting.

NeuroSynth also preserved stable Task A retention throughout continual training. Unlike EWC, however, NeuroSynth achieved improved retention while employing architectural consolidation mechanisms rather than relying exclusively on parameter regularization. The frozen hippocampal-inspired planning pathway may have contributed to the observed retention improvements during later adaptation.

After Task C training, NeuroSynth achieved a mean Task A success rate of 18.00\%, compared to 0.33\% for PPO. Two-sample statistical analysis demonstrated statistical significance ($p = 0.014929$) with a large effect size (Cohen's $d = 1.49$). These results indicate that NeuroSynth substantially reduced catastrophic forgetting relative to PPO while maintaining continued sequential learning capability.

\subsection{Task B Retention Across Continual Learning Phases}

\begin{figure}[ht]
    \centering
    \includegraphics[width=\textwidth]{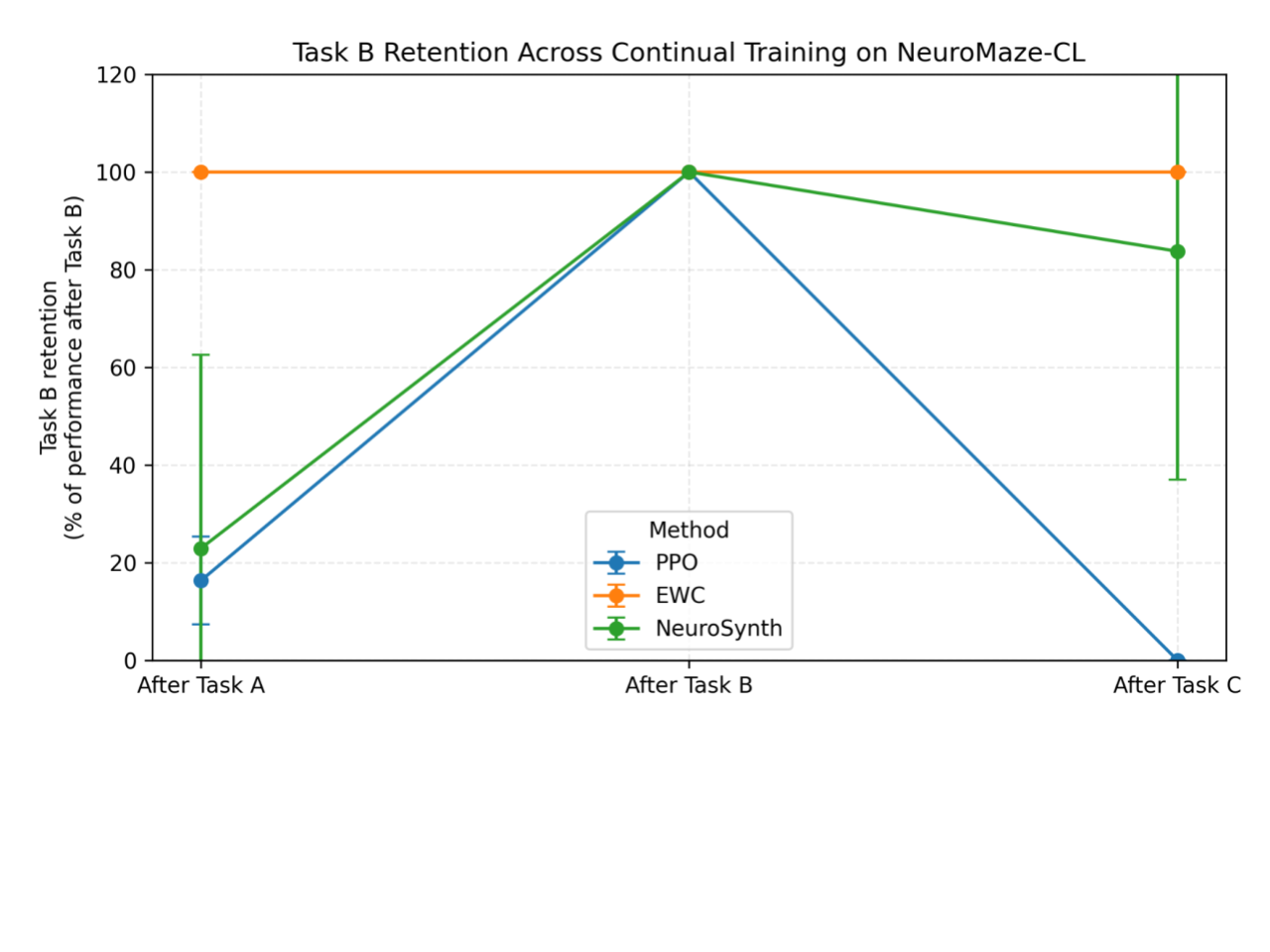}
    \caption{Task B Retention Across Continual Learning Phases. Task B success rate expressed as a percentage of post-Task B baseline performance, measured after each training phase. Error bars indicate one standard deviation across six seeds.}
    \label{fig:taskB}
\end{figure}

Figure~\ref{fig:taskB} shows Task B retention measured after each training phase. Because Task B was introduced after Task A, performance initially remained low during the earliest evaluation stage. After Task B training, all methods improved substantially on Task B performance. However, major differences emerged after Task C training.

PPO exhibited near-total forgetting of Task B after learning Task C, indicating that later-task gradients strongly interfered with intermediate task representations.

EWC maintained apparently stable Task B performance across all training phases. The consistency of EWC's Task B values across phases suggests that Fisher-based parameter constraints may have limited representational adaptation throughout the continual learning sequence.

NeuroSynth demonstrated substantially improved Task B success rate after Task C training. Following consolidation, NeuroSynth retained 35.33\% Task B performance compared to 0.00\% for PPO. Statistical testing demonstrated strong significance ($p = 0.002376$) with a very large effect size (Cohen's $d = 2.31$).

These findings suggest that NeuroSynth reduced interference between sequential tasks while preserving the ability to learn newly introduced environments.

\subsection{Task C Learning and Stability}

\begin{figure}[ht]
    \centering
    \includegraphics[width=\textwidth]{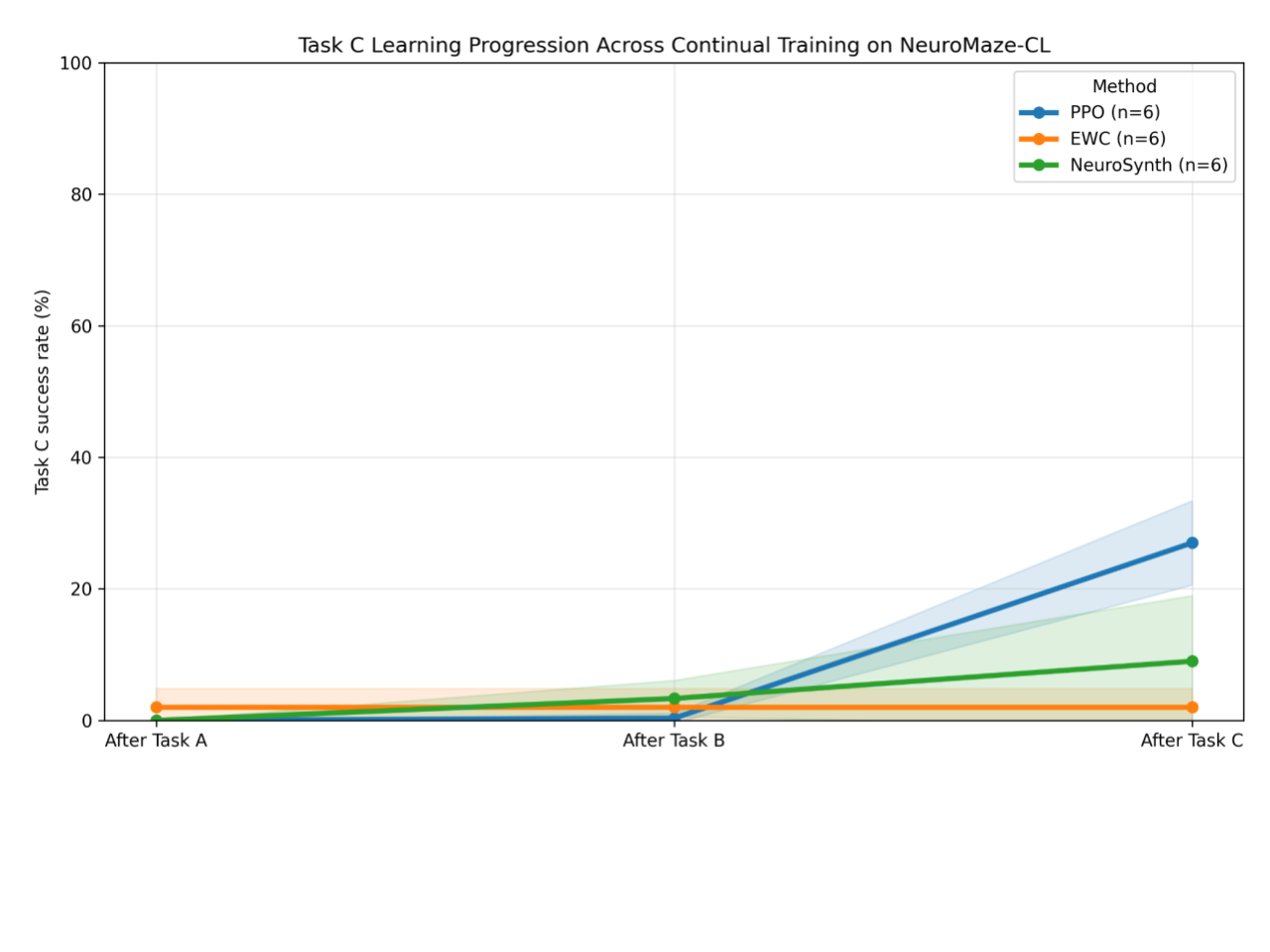}
    \caption{Task C Learning Progression Across Continual Training Phases. Task C success rate (\%) measured after each training phase across six seeds. Shaded regions indicate one standard deviation.}
    \label{fig:taskC}
\end{figure}

Figure~\ref{fig:taskC} visualizes Task C learning progression across continual-learning phases.

PPO demonstrated strong short-term Task C adaptation because unrestricted parameter updates allowed rapid optimization toward newly introduced tasks. However, this plasticity occurred at the expense of earlier-task retention.

EWC demonstrated comparatively weaker Task C acquisition. Fisher-based parameter constraints likely reduced the network's ability to substantially reorganize representations during later phases of continual learning.

NeuroSynth achieved improved Task C learning relative to EWC while simultaneously preserving earlier-task knowledge. Final Task C performance reached 9.00\% for NeuroSynth compared to 2.00\% for EWC. Although this difference did not achieve statistical significance ($p = 0.226643$), the observed moderate effect size (Cohen's $d = 0.56$) suggests meaningful performance improvements. Collectively, these results demonstrate that NeuroSynth achieved improved balance between long-term retention and continued adaptation relative to both baseline methods.

\FloatBarrier
\section{Discussion}

This study investigated whether biologically inspired memory consolidation mechanisms could improve continual reinforcement learning under sequential non-revisitation conditions. The results indicate that NeuroSynth reduced catastrophic forgetting relative to PPO while maintaining improved adaptability relative to EWC.

The PPO baseline demonstrated severe catastrophic forgetting after sequential training. Because PPO relies entirely on gradient-based optimization without replay or consolidation mechanisms, newly acquired representations overwrite previously learned behaviors. Under continual-learning conditions, this instability causes earlier task performance to collapse as the model adapts to later environments.

EWC partially mitigated forgetting through parameter regularization. By constraining updates to parameters estimated to be important for prior tasks, EWC preserved earlier representations more effectively than PPO. However, this stability came at the cost of reduced plasticity during later training phases. As tasks accumulated sequentially, parameter constraints increasingly limited the network's ability to reorganize representations for new environments.

NeuroSynth was designed to address the stability-plasticity tradeoff through architectural separation combined with replay and distillation mechanisms (5, 6, 12, 13). The hippocampal-inspired planning pathway was designed to support rapid acquisition and preservation of earlier-task knowledge, while the cortical-inspired habit pathway remained adaptive during future training phases. Replay and knowledge distillation mechanisms further reinforced retained representations throughout continual learning.

The observed retention improvements suggest that catastrophic forgetting may be mitigated through biologically inspired memory organization strategies that separate rapid acquisition from long-term consolidation. Rather than preserving knowledge exclusively through static parameter protection, NeuroSynth stabilized representations through modular memory pathways and replay-guided consolidation dynamics.

\subsection{Limitations}

Several limitations remain in the current implementation. First, experiments were conducted using relatively small grid-world navigation environments with discrete action spaces and low-dimensional observations. Although these environments effectively induce catastrophic forgetting, they do not fully represent the complexity of real-world reinforcement learning problems involving high-dimensional sensory inputs, continuous actions, stochastic dynamics, or long-horizon planning.

Second, the experimental setup involved only three sequential tasks. Larger continual-learning benchmarks containing many more tasks may produce different retention and stability behaviors. Additional large-scale evaluations are necessary to determine whether NeuroSynth scales effectively under prolonged lifelong learning conditions.

Third, NeuroSynth was evaluated primarily against PPO and EWC baselines. Although these baselines represent widely studied reinforcement-learning and continual-learning methods, additional comparisons against replay-based and hybrid continual-learning systems would strengthen the generalizability of the results (17, 18). Additionally, this study did not include component-level ablation experiments isolating the individual contributions of replay, pathway separation, and distillation mechanisms.

Finally, the current implementation remains biologically inspired rather than biologically realistic. The hippocampal-cortical abstraction used in NeuroSynth simplifies many known biological consolidation processes, including sleep-dependent replay, synaptic plasticity modulation, and distributed cortical integration.

\subsection{Future Work}

Several important directions remain for future development of NeuroSynth. First, future research should evaluate NeuroSynth in large-scale continuous-control and robotic environments. The current experiments were conducted in grid-world navigation tasks with discrete actions and relatively low-dimensional observations. Real-world robotic systems must continuously adapt to environmental changes while preserving previously learned motor behaviors, making continual reinforcement learning particularly important. Applying NeuroSynth to robotic manipulation, autonomous navigation, embodied control systems, or drone coordination tasks would provide stronger evidence regarding whether biologically inspired consolidation mechanisms can scale to realistic lifelong-learning problems.

Second, future work should investigate NeuroSynth under sparse-reward and procedurally generated environments. Sparse-reward settings create substantial exploration challenges because successful trajectories occur infrequently, increasing the difficulty of preserving useful behaviors across sequential tasks. Under these conditions, catastrophic forgetting may become more severe due to the rarity of informative experiences. Evaluating NeuroSynth in procedurally generated environments would additionally test whether the architecture can generalize across changing task structures rather than memorizing fixed layouts or deterministic navigation patterns. Indeed, future studies should perform ablation analyses to quantify the relative contribution of each architectural component to retention and adaptation performance.

Third, future implementations should explore adaptive consolidation mechanisms instead of static pathway freezing. In the current architecture, the hippocampal-inspired planning pathway is permanently frozen following Task A training. Although this improves retention, permanently freezing representations may reduce long-term flexibility as task complexity increases. Future architectures could investigate dynamic consolidation schedules, selective reactivation of frozen pathways, or meta-learned replay policies that continuously balance stability and plasticity during lifelong learning.

Finally, future studies should evaluate NeuroSynth against broader continual-learning benchmark suites and additional baseline methods, including replay-based continual reinforcement-learning systems and transformer-based memory architectures. Larger statistical populations and more complex task distributions would improve the reliability and scalability of the current findings.

\FloatBarrier
\section{Conclusion}

This study introduced NeuroSynth, a biologically inspired continual reinforcement learning architecture designed to reduce catastrophic forgetting through dual-pathway memory consolidation, replay stabilization, and knowledge distillation mechanisms. Experimental results demonstrated that PPO suffers from severe catastrophic forgetting under sequential non-revisitation conditions because newly acquired gradients overwrite previously learned representations. EWC partially mitigated forgetting through parameter regularization but demonstrated reduced adaptability during later tasks due to increasingly restrictive parameter constraints.

NeuroSynth addressed these limitations by reorganizing memory storage into separate acquisition and consolidation pathways inspired by hippocampal-cortical learning systems. Across sequential navigation tasks, NeuroSynth preserved previously learned behaviors while maintaining continued adaptability to new environments, outperforming PPO and demonstrating moderate improvements relative to EWC. These findings suggest that architectural design may play an important role in mitigating catastrophic forgetting under continual reinforcement learning conditions. Biologically inspired consolidation mechanisms may therefore provide a promising direction for the development of stable lifelong reinforcement learning systems.

\section*{Acknowledgements}

The author would like to sincerely thank Dr. Vineet Shenoy for his guidance, mentorship, and support throughout this research. The author is also grateful to the Aspiring Scientists Summer Internship Program (ASSIP) at George Mason University for providing the opportunity and collaborative environment to pursue interdisciplinary research at the intersection of artificial intelligence, neuroscience, and continual reinforcement learning. Finally, the author acknowledges the developers of the open-source frameworks and tools, including MiniGrid and PyTorch, that supported this work. Disclaimer. This manuscript is provided as an arXiv preprint to establish a public record of the NeuroSynth continual reinforcement learning architecture. A shortened applications-focused version evaluating NeuroSynth in an ICU sepsis setting has been prepared for submission to the Columbia Junior Science Journal (CJSJ). This full manuscript has also been submitted to the Journal of High School Science for peer review.

\bibliographystyle{unsrt}

\section*{Code Availability}

The complete NeuroSynth implementation, NeuroMaze-CL benchmark, baseline implementations (PPO and EWC), evaluation scripts, and figure generation code are publicly available at

\begin{center}
\url{https://github.com/yashkini1/NEUROSYNTH}
\end{center}

\appendix
\section{Appendix}

\begin{figure}[ht]
    \centering
    \includegraphics[width=\textwidth]{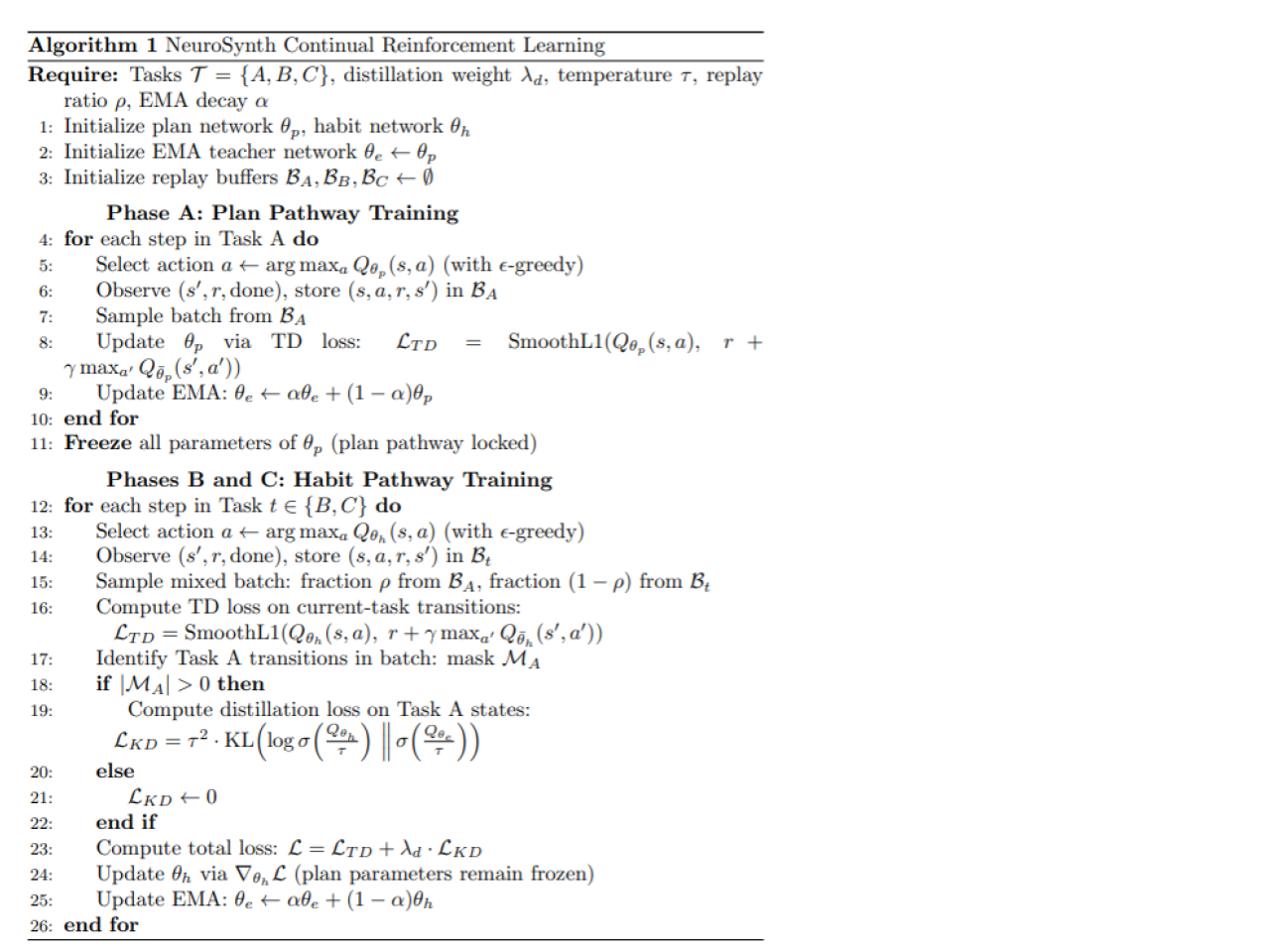}
    \caption{Pseudo-code for the NeuroSynth continual reinforcement learning algorithm. The planner pathway is trained on Task A and subsequently frozen, while the habit pathway learns Tasks B and C using replay-based rehearsal and KL-divergence policy distillation from the frozen planner.}
    \label{alg:neurosynth}
\end{figure}

\end{document}